\documentclass[11pt]{article}

\usepackage[final]{acl}

\usepackage{times}
\usepackage{latexsym}

\usepackage[T1]{fontenc}

\usepackage[utf8]{inputenc}

\usepackage{microtype}

\usepackage{inconsolata}

\usepackage{graphicx}

\usepackage[vlined,ruled]{algorithm2e}
\usepackage{makecell}
\usepackage{multirow}
\usepackage{subcaption}
\usepackage[many]{tcolorbox}
\usepackage{tikz}
\usetikzlibrary{positioning, shapes, arrows.meta}
\usepackage{adjustbox}
\usepackage{booktabs}
\usepackage{amssymb}

\title{Out-of-Distribution Federated Distillation with Domain-Aware Proxy}

\author{
Jiahao Xiao
\and
Jiangming Liu\textsuperscript{*}
\\
School of Information Science and Engineering,
Yunnan University, China
\\
\textsuperscript{*}Corresponding author: jiangmingliu@ynu.edu.cn
}

\begin{document}
\maketitle
\begin{abstract}
Federated Learning is a distributed machine learning paradigm that trains a global model by aggregating local clients without sharing private data of each client. Federated Distillation (FD) builds upon this paradigm by leveraging knowledge distillation to exchange soft predictions on proxy data instead of model parameters, enabling more efficient communication and supporting heterogeneous model collaboration. However, FD models trained on In-Distribution data are hardly adapted to Out-of-Distribution (OOD) scenarios. In this paper, we propose a domain-aware proxy selection framework to better adopt proxy data for OOD problems. The experimental results show that the proposed models effectively address the challenges of distribution shifts under OOD with and without proxy data by achieving average 82.9\% and 80.6\% over existing works on standard benchmarks. The codes and data are released in~\url{https://anonymous.4open.science/r/DPS-FD-8596/}.
\end{abstract}

\section{Introduction}
Federated Learning is an emerging distributed machine learning paradigm that trains a global model by aggregating local clients without sharing private data kept by each client~\cite{mcmahan2017communication,abadeer2022flightner,pei2024review,Chen_2024_CVPR,Wang_2025_CVPR}. However, standard Federated Learning has the limitations of the high communication cost and the homogeneous customization.
To address these limitations, Federated Distillation (FD) introduces shared proxy data as a medium for knowledge transfer between clients and the server~\cite{hinton2015distilling,you2017learning,li2019fedmd}. By exchanging soft labels on proxy data instead of model parameters, FD significantly reduces communication overhead and enables the use of heterogeneous models across clients~\cite{lin2020ensemble,hu2021mhat,Wu_2022,Itahara_2023,shao2024selective,fan2025ten,wang2025fed}. 

However, FD models trained on In-Distribution data are hardly adapted to Out-of-Distribution (OOD) scenarios, due to the proxy data that serves as a critical medium for knowledge transfer~\cite{hendrycks2016baseline,pmlr-v202-guo23b,bai2023feed,10687423,qi2025federated,jeong2025out}. 
Since proxy data heavily affect distribution shifts under OOD, we propose $\textbf{DPS-FD}$, a $\textbf{D}$omain-aware $\textbf{P}$roxy $\textbf{S}$election framework for $\textbf{FD}$ to better adopt proxy data for OOD problems. Specifically, DPS-FD enables each local client to select the relevant domain samples, while the server selects globally-representative samples from the proxy data. By taking the union of these selected samples, DPS-FD constructs a domain-aware proxy data that better captures the characteristics of both local and global distributions.

Although the effective adoption of proxy data can mitigate OOD promblem in FD, the original proxy data sometimes cannot be obtained for knowledge transfer in FD~\cite{ZHOU2023120327,10.1093/nsr/nwae276,fang2024data}. These works mainly focus on computer visions~\cite{takahashi2023breaching,liao2024foogd,qi2025federated,wang2025diffusionmodelbaseddatasynthesis}. 
To address the challenge of no proxy data in textual data, we propose a \textbf{V}ocabulary-\textbf{C}onstrained LLM-based generation (\textbf{VC}) strategy for generating proxy data, where a global vocabulary is adopted as lexical constraints to guide the proxy data generation. VC adopts a few-shot prompting strategy that consists of two complementary prompt templates: system prompts and user prompts, to generate diverse and high-quality proxy data. According to the comprehensive empirical experiments, we find that high-quality proxy data can significantly mitigate OOD problems. DPS-FD generates proxy data using the VC strategy if the proxy data is unavailable.




Experiment results on standard benchmarks demonstrate that DPS-FD achieves competitive performance both with and without proxy data. Additionally, we take insight analysis on the effect of the high-quality diverse proxy data, unveiling that global distribution of the proxy data heavily affect the OOD performance of the global model in server. The main contribution of this paper are as follows:
\vspace{-0.5em}
\begin{itemize}
\item We propose a domain-aware proxy selection framework that better adopts proxy data for OOD problems, enhancing the robustness and generalization of the global model.
\vspace{-0.5em}
\item We introduce a vocabulary-constrained LLM-based generation strategy to enable FD models to maintain competitive performance without access to real proxy data.
\vspace{-0.5em}
\item We conduct analysis on the role of proxy data in FD, revealing that high-quality proxy data are crucial for mitigating OOD problems and enhancing the robustness of the global model.
\end{itemize}

\section{Related Work}
\subsection{Federated Learning and Distillation}
FL emerges as a promising distributed learning paradigm that enables collaborative model training without sharing private data of each client. Classic approaches such as FedAvg~\cite{mcmahan2017communication} aggregate model parameters from clients into a global model and iteratively repeats this process. However, they suffer from high communication overhead, require homogeneous model architectures, and privacy-leakage across clients and the server~\cite{sui-etal-2020-feded,lin2021ensembledistillationrobustmodel}.
To mitigate these limitations, FD leverages knowledge distillation~\cite{hinton2015distilling,you2017learning,anil2018large} to exchange soft labels on shared proxy data instead of model parameters~\cite{Wu_2022,Itahara_2023,Chen_2024_CVPR,wang2025fed}, significantly reducing communication costs, and enabling heterogeneous models.

\subsection{OOD in Federated Distillation}
OOD behavior manifests primarily as covariate shifts and semantic shifts~\cite{liao2024foogd}, which often leads to model degradation in FD~\cite{gulrajani2020search}. To address this issue, FD extends to OOD scenarios by enhancing the alignment between proxy and client distributions~\cite{yu2023turning,qi2025federated}. Existing work leverages public or synthesized proxy data to reduce domain gaps and improve knowledge transfer~\cite{jeong2023communicationefficientondevicemachinelearning}. \citet{zhu2021data} introduce adaptive weighting and ensemble strategies to emphasize client-specific contributions during distillation and enhance robustness. These developments highlight FD potential in improving generalization in OOD environments.
To further advance this line of research, we propose DPS-FD, which constructs high-quality proxy data to model the diversity across clients, enabling effective and efficient knowledge transfer.

\subsection{Proxy Data in Federated Distillation}
Proxy data plays a crucial role in FD, serving as the essential medium through which knowledge is exchanged between heterogeneous clients and the global model~\cite{li2019fedmd,lin2020ensemble}. Without proxy data, the distillation process struggles to align knowledge~\cite{liao2023joint,liao2024rethinking,xiao2025adaptive}.

To address these limitations, existing studies explore various generative approaches to synthesize proxy data for FD. These include 1) logit-based inversion methods, which reconstruct pseudo data by optimizing inputs to match client logits~\cite{takahashi2023breaching}; 2) generator-based approaches using GANs or VAEs to learn data distributions and produce representative samples~\cite{zhang2022fedzkt,wang2023dafkd,liao2024foogd,ma2025data,qi2025federated}; 3) diffusion-based models, which iteratively denoise random noise into high-quality synthetic data~\cite{li2023syntheticdatadiffusionmodels,wang2024datafreefederatedclassincremental,yang2024feddeodescriptionenhancedoneshotfederated,wang2025diffusionmodelbaseddatasynthesis}.

However, in natural language processing, the situation becomes more challenging. Most existing proxy-based distillation methods are designed for vision tasks and overlook fundamental linguistic properties such as lexical frequency distribution and semantic structure. This gap highlights the need for domain-aware proxy construction methods tailored to textual data and motivates our work on leveraging lexical constraints and large language models to build more effective proxy data in FD.

\begin{figure*}[!tp]
    \centering
    \includegraphics[scale=0.5]{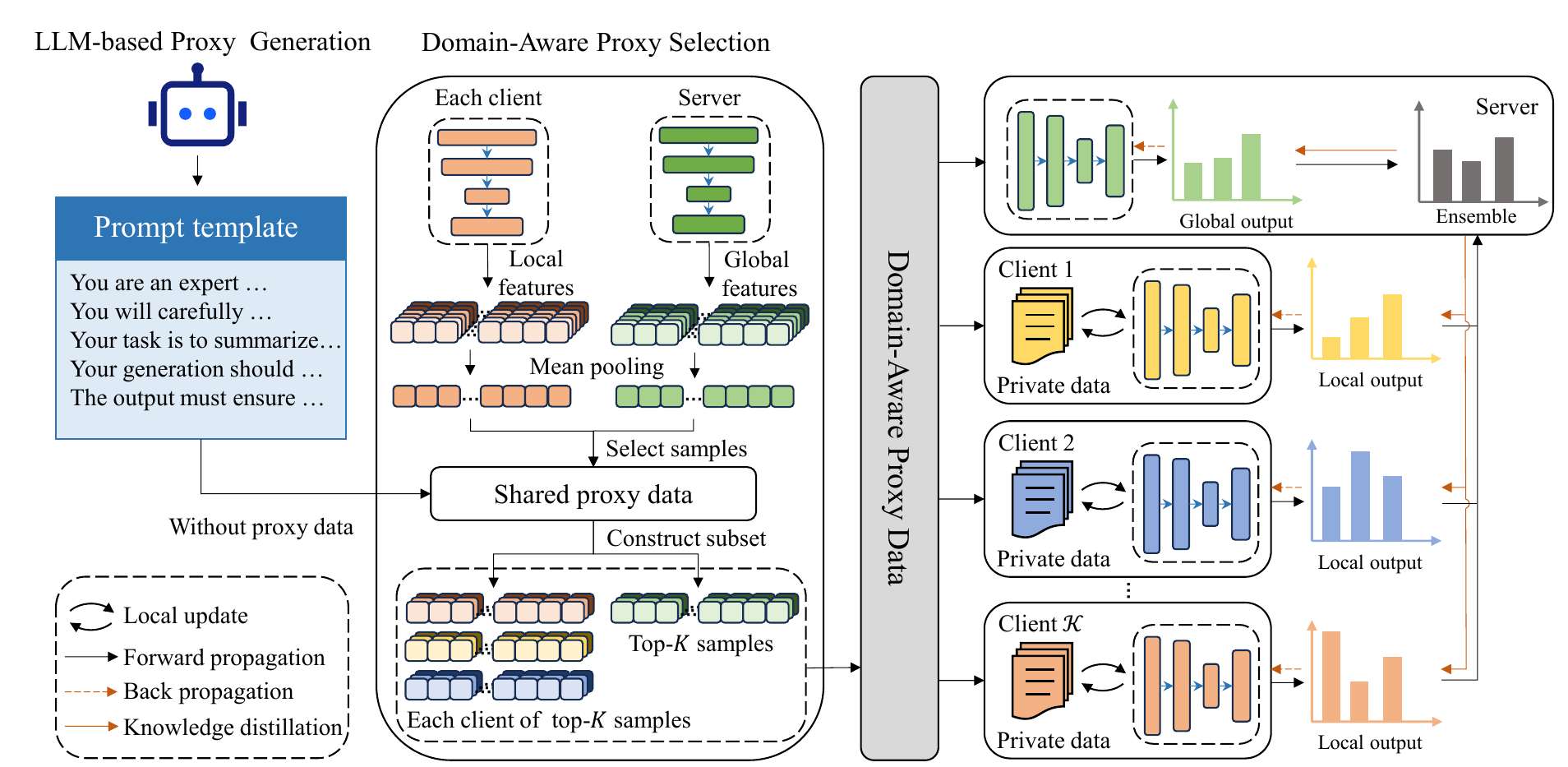}
    \caption{DPS-FD.}
    \label{fig:dps-fd}
\end{figure*}

\section{Preliminary}
We take classification as an example to investigate OOD problems in FD. In the context of a classification task under FL with $K$ clients and a central server, each client $k\in\{i=1,2,\dots,K\}$ locally keeps its own private labeled data $D_k = \{(x_i^k,y_i^k)\}_{i=1}^{N_k}$ that is isolated from others, where $x_i^k \in \mathbb{R}^d$ and $y_i^k \in \{1,2,\dots,C\}$ are the $i$-th instance and its corresponding label, respectively, $C$ is the number of classes, and $N_k$ is the size of private data. The objective is to train a globally optimal model $f_g$ parameterized by $\theta_g$ through aggregating client models $f_k$ parameterized by $\theta_k$ in a decentralized manner without exposing private data on the server.

To mitigate the limitations of FL based on parameter exchange, FD leverages knowledge distillation to exchange soft labels instead of model parameters by introducing shared proxy data $D_p = \{(x_i^p)\}_{i=1}^{N_p}$ for knowledge transfer between clients and the server, where $N_p$ is the size of the proxy data. Specifically, in each communication round $t \in T$, the server randomly selects the set of activated clients $\mathcal{K} \subseteq \{1,2,\dots,K\} $ based on a sampling fraction $\epsilon$ to participate in the FD training, where $|\mathcal{K}|=|\epsilon \cdot K|$. Each client $\mathcal{K}$ first trains its local model on its own private data $D_k$, and the local optimization objective is defined as follows:
\begin{equation}
    \theta^*_k=\arg\min_{\theta_k} \ \mathbb{E}_{(x,y) \sim D_k}[\mathcal{L}_{\text{CE}}(f_k(x;\theta_k),y)],
    \label{eq:local-train}
\end{equation}
where $\mathcal{L}_{\text{CE}}(\cdot)$ denotes the cross-entropy loss used for local classification tasks, and $\mathbb{E}$ denotes its expected value with respect to the local data distribution $D_k$. The participating clients compute soft labels on the proxy data $D_p$ and upload them to the server. The server then aggregates these labels using a weighting scheme proportional to the size of each client local data to distill the global model. The corresponding objective function is formulated as:
\begin{equation}
    w_k=\frac{N_k}{||N||_1},~N=[N_1,N_2,\dots,N_{\mathcal{K}}],
    \label{eq:server-agg}
\end{equation}
\begin{equation}
    h(x_p) = \sigma(f(x_p;\theta)),
\end{equation}
\begin{equation}
\small
\begin{aligned}
    \theta_g^* = \arg\min_{\theta_g} \
    \mathop{\mathbb{E}}\limits_{x_p \sim D_p}
    \big[
        \mathcal{L}_{\text{KL}}\big(
        \sum_{k \in \mathcal{K}} w_k \cdot h_k(x_p),
        \ h_g(x_p)
        \big)
    \big],
    \label{eq:server-distill}
\end{aligned}
\end{equation}
where $\sigma(\cdot)$ is the softmax function with temperature $\tau$ to control the smoothness of soft labels, and $\mathcal{L}_{\text{KL}}$ is the Kullback-Leibler divergence used to measure the distribution differences between the server and the aggregated local models. Finally, each client further distills its local model using the KL divergence with the global soft labels predicted by the global model on the proxy data $D_p$.

\begin{equation}
    \theta_k^*=\arg\min_{\theta_k} \ \mathbb{E}_{x_p \sim D_p}
    \big[
        \mathcal{L}_{\text{KL}}\big(
        h_g(x_p), h_k(x_p)
        \big)
    \big].
    \label{eq:client-distill}
\end{equation}

\section{Method}
We construct OOD benchmarks in natural language processing tasks to investigate the challenges of FD in real-world scenarios. Then, we propose \textbf{DPS-FD}, a novel framework of \textbf{D}omain-aware \textbf{P}roxy \textbf{S}election for FD to address the OOD challenge by reconstructing the domain-aware proxy data.
Furthermore, we introduce a vocabulary-constrained LLM-based proxy generation strategy coupled with the DPS mechanism to mitigate the OOD problems of FD without proxy data.
\subsection{OOD Benchmark Construction}
To better understand the challenges of FD in real-world scenarios, we investigate the role of proxy data in OOD settings. Before introducing the OOD setting, we first consider a multi-domain non-IID configuration in natural language processing, which accounts for both label distribution skew and the domain distribution of each client~\cite{xiao2025adaptive,yang2023personalized,xiao2024confusion,mao2025fedkt,yan2025fedvck}. Specifically, each client keeps private data from a distinct domain, and the label distribution across clients is made heterogeneous by sampling according to a Dirichlet distribution with different $\alpha$ parameters. 
Building upon the multi-domain non-IID setup, we simulate OOD settings by replacing the original test set, which is composed of data from the client domains, with a test set sampled from the domains not seen by the clients.
Both the private data and proxy data are sampled from an open-source Amazon product review database.\footnote{https://archive.org/details/amazon-reviews-1995-2013.} The domain-level feature distributions are visualized in Appendix~\ref{appendix:domain-diversity}.

\subsection{Domain-Aware Proxy Selection}
As shown in Figure~\ref{fig:dps-fd}, in communication round $t$, each client optimizes its local model using its labeled private data and utilizes the DPS mechanism to reconstruct domain-aware proxy data and uploads the corresponding local outputs to the server.
The server aggregates the outputs and optimizes the global model through knowledge distillation. The global outputs given by the server are subsequently broadcast to all clients for local distillation, enabling bidirectional knowledge transfer.

DPS enables each client $k$ to select the most relevant domain samples from the proxy data $D_p$, allowing clients to transmit high-confidence, domain-specific knowledge that enhances the robustness of the global model, while the server selects globally representative samples to improve the generalization of the global model. By taking the union of these selected samples, we construct a domain-aware proxy data $D_p^*$ that serves as a more effective and efficient knowledge transfer medium between clients and the server.


In each communication round $t \in \{1, \dots, T\}$, after local training on its private data $D_k$, client $k$ computes the centroid representation $c_k$ of its domain, and the server computes the global centroid $c_g$ from the proxy data $D_P$ as:
\begin{equation}
\begin{aligned}
    & \mathbf{c}_k = \frac{1}{{N_k}} \sum_{(x, y) \in D_k} f_k(x;\theta_k),
    \\
    & \mathbf{c}_g = \frac{1}{{N_p}} \sum_{x \in D_p} f_g(x;\theta_g),
    \label{eq:centroid}
\end{aligned}
\end{equation}
where $f_k(x)$ and $f_g(x)$ denote the input \(x\) feature representations of the local model and the global model, respectively.
For each proxy sample $x_i \in D_P$, we calculate its cosine similarity with each client centroid and the global centroid:
\begin{equation}
\begin{aligned}
    & \mathrm{sim}_k(x_i) = \frac{f_k(x_i;\theta_k) \cdot \mathbf{c}_k}{\|f_k(x_i;\theta_k)\| \, \|\mathbf{c}_k\|}, 
    \\
    & \mathrm{sim}_g(x_i) = \frac{f_g(x_i;\theta_g) \cdot \mathbf{c}_g}{\|f_g(x_i;\theta_g)\| \, \|\mathbf{c}_g\|}.
    \label{eq:proxy-sims}
\end{aligned}
\end{equation}

Each client and the server then select the top-\(K\) proxy samples most similar to its domain representations, respectively:
\begin{equation}
\begin{aligned}
    P_k = \mathrm{TopK}\bigl( \{ x_i \in D_P \mid \mathrm{sim}_k(x_i) \} ),\\
    P_g = \mathrm{TopK}\bigl( \{ x_i \in D_P \mid \mathrm{sim}_g(x_i) \} ).
    \label{eq:proxy-topk}
\end{aligned}
\end{equation}
Finally, the domain-aware proxy data used for knowledge distillation is constructed as the union:
\begin{equation}
    D_p^{*} = \bigcup_{k=1}^{\mathcal{K}} P_k \cup P_g.
    \label{eq:proxy-union}
\end{equation}
This selection process is repeated in each communication round, ensuring that the proxy data dynamically aligns with evolving domain representations as the training progresses. DPS-FD algorithm is given in Appendix~\ref{appendix:algo}.

\subsection{LLM-Based Proxy Generation}
The global model benefits from additional knowledge that helps align the client-specific distributions and facilitates more effective knowledge transfer. High-quality proxy data provides a bridge for the global model to capture patterns not fully represented in individual client private data, enhancing both robustness to domain-specific variations and generalization to unseen domains.
In the absence of proxy data, the global model has to rely solely on local clients, leading to weak alignment among local distributions.

\begin{figure}[t]        
  \centering
  \scalebox{0.5}{\includegraphics{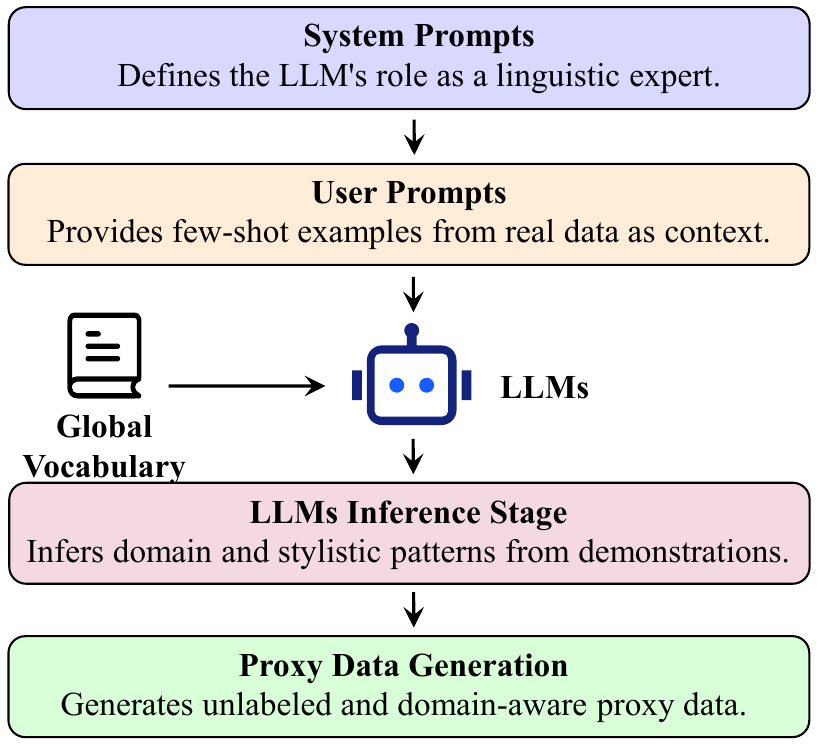}} 
  \caption{LLM-based Proxy Generation with Vocabulary-Constraints.}
  \label{fig:llm-proxy}
\end{figure}

To address the challenge of unavailable proxy data in FD, we propose a vocabulary-constrained LLM-based proxy data generation strategy. As shown in Figure~\ref{fig:llm-proxy}, we first construct a vocabulary from tokenizing real proxy data, which serves as a constraint to ensure that generated samples adhere to realistic lexical distributions and domain-specific semantics.
To effectively generate diverse and high-quality proxy data, we adopt a few-shot prompting strategy, consisting of two prompts:
\begin{itemize}
\vspace{-0.7em}
    \item System prompts define the overall role and objective of the LLM, instructing it to act as a linguistic expert specialized in generating diverse, fluent, and contextually coherent product review texts tailored for FL tasks.
    \vspace{-0.7em}
    \item User prompts provide few-shot representative samples drawn from real environments as in-context demonstrations.
\end{itemize}
To ensure linguistic consistency and data complexity, we constrain the average sentence length to approximately 90 words per generated text. As shown in Figure~\ref{fig:llm-proxy}, two-stage prompting design allows the LLM to capture both global generation intent and domain-specific semantics.
\footnote{The detailed structures of the system and user prompt are given in Appendix~\ref{appendix:prompt}.}


\section{Experiments}
We evaluate our proposed methods, DPS-FD along with a vocabulary-constrained LLM-based generation under OOD both with and without proxy data.

\subsection{Settings}
We conduct experiments on a sentiment classification task under heterogeneous client settings, considering both the presence and absence of proxy data. To simulate OOD scenarios, data are sampled by label within each domain from the Dirichlet distribution with concentration parameters $\alpha=\{0.3, 0.5, 0.8, 1.2, 1.5 \}$ for clients, respectively, to control the degree of heterogeneity.
\vspace{-1em}
\paragraph{Backbones} We assign different pre-trained language models to the 5 clients, namely \texttt{BERT-base-cased}, \texttt{BERT-large-cased}, \texttt{RoBERTa-base}, \texttt{RoBERTa-large}, and \texttt{XLNet-large-cased} model, respectively, while using \texttt{RoBERTa-large} as the server model.\footnote{All the pre-trained language model cards are sourced from \url{https://huggingface.co/models}.}

\vspace{-1em}
\paragraph{Data} Clients keep local data from specific domains, i.e., automotive (a.), baby (b.), clothing (c.), health (h.), and sport (s.), respectively, while the global distribution (g.) consists of all domains unseen by the clients.

\vspace{-1em}
\paragraph{Metrics}
We use F1 scores to evaluate classification performance, cost to measure communication efficiency during knowledge transfer, and ECE together with prediction entropy to assess model calibration and uncertainty analysis.

\begin{table}[!tp]
    \vspace{-2mm} 
    \setlength{\tabcolsep}{3pt}  
    \resizebox{\columnwidth}{!}{
    \begin{tabular}{l|c|ccccc|c|c|c}
    \toprule
    Model & \textbf{g.} & \textbf{a.} & \textbf{b.} & \textbf{c.} & \textbf{h.} & \textbf{s.} & cost (\%) & ECE & entropy  \\
    \midrule
    \multicolumn{10}{c}{with proxy data} \\
    \midrule
    FD-LP & 80.7 & 79.8 & 84.7 & 91.4 & 77.4 & 81.8 & 100 & 13.89 & 0.18 \\
    FD-GP & 81.4 & 80.3 & 84.1 & 85.0 & 75.6 & 78.7 & 100 & 7.03 & 0.33  \\
    \midrule
    DPS-FD & \textbf{82.9} & 80.6 & 85.1 & 81.9 & 76.0 & 78.9 & 98.3 & 9.84 & 0.20 \\
    \midrule
    \multicolumn{10}{c}{without proxy data} \\
    \midrule
    FD-VC & 78.2 & 75.0 & 80.5 & 73.9 & 68.4 & 72.7 & 100 & 16.58 & 0.18  \\
    \midrule
    DPS-FD & \textbf{80.6} & 77.5 & 81.7 & 74.6 & 73.5 & 76.3 & 95.9 & 8.68 & 0.29  \\
    ~~~~w/o VC & 77.7 & 75.8 & 79.3 & 77.1 & 66.6 & 71.6 & 93.9 & 7.00 &  0.40 \\

    \bottomrule
    \end{tabular}}
    \caption{Results of the global model in heterogeneous OOD settings  ($p < 0.001$). Column \textbf{g.} is score on test data in global distribution, while columns \textbf{a.}, \textbf{b.}, \textbf{c.}, \textbf{h.} and \textbf{s.} are scores on test data in respective domains.}
    \label{tab:main-results}
\end{table}

\subsection{Training and Inference}
We train the models for 5 communication rounds with 3 local epochs per round, using an initial learning rate of 2e-5. AdamW is adopted as the optimizer, with a maximum sequence length of 128 and a batch size of 32. The number of clients is set to 5, corresponding to different numbers of domain-specific data. For the proposed DPS-FD method, we select the top 25k samples on each client and the top 35k samples on the server to form the refined proxy data. GPT-3.5-turbo is used as the LLM in the LLM-based generation strategy. We use an NVIDIA GPU A100 for training and inference.

\begin{table}[!tp]
\vspace{-2mm} 
    \small
    \resizebox{\columnwidth}{!}{
    \begin{tabular}{l|ccccc}
    \toprule
    Model & \textbf{a.} & \textbf{b.} & \textbf{c.} & \textbf{h.} & \textbf{s.} \\
    \midrule
        \multicolumn{6}{c}{with proxy data} \\
        \midrule
    FD-LP & 78.2 & 47.6 & 93.4 & 76.9 & 81.5 \\
    FD-GP & 78.0 & 47.6 & 93.2 & 76.0 & 81.4 \\
    \midrule
    DPS-FD & 78.5 & 84.6 & 90.4 & 76.5 & 82.3 \\
    \midrule
            \multicolumn{6}{c}{without proxy data} \\
            \midrule
    FD-VC & 76.3 & 47.6 & 94.0 & 73.3 & 77.7 \\
    \midrule
     DPS-FD & 76.9 & 81.0 & 89.8 & 67.7 & 77.1 \\
    ~~~~w/o VC & 75.2 & 82.1 & 91.3 & 70.0 & 72.8 \\

    \bottomrule
    \end{tabular}}
    \caption{Results of local models on clients ($p < 0.001$). The columns \textbf{a.}, \textbf{b.}, \textbf{c.}, \textbf{h.} and \textbf{s.} are scores on test data in respective domains.}
    \label{tab:client-results}
\end{table}

\subsection{Baselines}
We evaluate the proposed DPS-FD against a series of baselines to thoroughly investigate the impact of proxy data and the effectiveness of our approach under OOD settings.
\begin{itemize}
\vspace{-0.7em}
    \item \textbf{FD-LP} refers to the standard FD method~\cite{lin2020ensemble}, which uses proxy data consisting only of samples from the client domains.
    \vspace{-0.7em}
    \item \textbf{FD-GP} uses global proxy data composed of samples from all domains that are unseen by the clients.
    \vspace{-0.7em}
    \item \textbf{FD-VC} leverages a vocabulary-constrained LLM-based generation strategy to synthesize proxy data without proxy data.
    \vspace{-0.7em}
    \item \textbf{DPS-FD} is our proposed model, applying DPS on FD with global domain proxy data, with proxy data generated by a vocabulary-constrained LLM, and with proxy data generated by an unconstrained LLM.
\end{itemize}

\subsection{Results}
Table~\ref{tab:main-results} and ~\ref{tab:client-results} show the results of the global server model and the local client models, respectively, on our proposed model and several baseline models.

\vspace{-0.5em}
\paragraph{Proxy Data Influence}

As shown in Table~\ref{tab:main-results}, the global model of FD-GP outperforms FD-LP, indicating that global-domain proxy data, compared to local-domain proxy data, can more effectively mitigate distribution shifts under OOD settings.

With proxy data, our proposed DPS-FD achieves superior performance across both global and local domains except for the clothing domain. By incorporating DPS, clients and the server dynamically identify and utilize proxy samples that better align with the distributions of local and global domains. DPS effectively filters out noisy or irrelevant samples, leading to high-quality proxy data that enable more effective and efficient knowledge transfer.
Without proxy data, DPS-FD outperforms DPS-FD without vocabulary-constraint by approximately 2.9\%, demonstrating that vocabulary-constrained LLM generation produces higher-quality proxy data. Moreover, DPS-FD substantially outperforms FD-VC, remaining effective even with synthetic proxy data.
\vspace{-0.5em}
\paragraph{Calibration and Uncertainty Estimation}

As shown in Table~\ref{tab:main-results}, with proxy data, replacing local-domain proxy data (FD-LP) with global-domain proxy data (FD-GP) significantly improves robustness under distribution shifts, resulting in lower ECE and higher entropy. Although DPS-FD does not achieve the lowest ECE, it maintains balanced calibration with moderate entropy, avoiding overconfident predictions.
Without proxy data, FD-VC exhibits severe overconfidence, while incorporating DPS improves robustness with lower ECE and higher entropy. Although DPS-FD without VC achieves the lowest ECE, the corresponding performance drop suggests that excessive diversity may introduce noisy samples, suggesting the benefit of vocabulary-constrained generation for balancing performance and robustness.

\begin{table}[!tp]
\vspace{-2mm} 
\centering
    
    \resizebox{\columnwidth}{!}{   
    \begin{tabular}{lccccc}
    \toprule
    DPS-FD & top-$K$ (client, server) & \textbf{g.} & cost (\%) & ECE & entropy\\
    \midrule
    \multicolumn{6}{c}{5 Clients} \\
    \midrule
    \multirow{2}{*}{~~ w/ proxy data}
     & (15k, 15k) & 82.6 & 89.8 & 9.68 & 0.22 \\
     & (25k, 35k) & \textbf{82.9} & 98.3 & 9.84 & 0.20  \\
    \multirow{2}{*}{~~ w/o proxy data}
     & (15k, 15k) & 77.5 & 78.3 & 13.68 & 0.26 \\
     & (25k, 35k) & 80.6 & 96.3 & 8.68 &  0.29 \\
    \midrule
    \multicolumn{6}{c}{10 Clients} \\
    \midrule
    \multirow{2}{*}{~~ w/ proxy data}
     & (15k, 15k) & \textbf{84.2} & 95.9 & 7.36 & 0.22 \\
     & (25k, 35k) & 84.0 & 99.8 & 11.69 &  0.13 \\
    \multirow{2}{*}{~~ w/o proxy data}
     & (15k, 15k) & 81.4 & 80.3 & 10.97 & 0.20 \\
     & (25k, 35k) & 82.7 & 93.6 & 5.86 & 0.28 \\
    \bottomrule
    \end{tabular}}
    \caption{Results of DPS-FD across different number of clients and top-$K$ selections.}
    \label{tab:dps_nclient_topk}
\end{table}
\vspace{-0.5em}
\paragraph{Local Models}
Table~\ref{tab:client-results} shows F1 scores of each local model evaluated on their own local test set. Existing methods perform poorly on local distribution of the baby domain, achieving only 47.6\% F1 scores, whereas our proposed DPS-FD substantially achieves the improvements on the local test set. DPS effectively mitigates the interference from other domains and improves the performance of local models within their own domains by filtering out noise and irrelevant proxy samples. This demonstrates that DPS-FD not only enhances the global model on server but also somehow enhances the local model on clients. 
\vspace{-0.5em}
\paragraph{Communication Cost}
To facilitate a fair comparison, we normalize the communication cost of the traditional FD method to 100\% as a baseline. As shown in the column cost of Table~\ref{tab:main-results}, DPS-FD achieves a slight yet meaningful reduction in communication cost from 100\% to 98.3\% with proxy data and from 100\% to 95.9\% without proxy data. The overall cost is determined by the number of clients $\mathcal{K}$, the proxy data size $|D_P|$, and the number of communication rounds $T$, and is typically expressed as $(\mathcal{K}+1) \cdot |D_P| \cdot T$. By constructing a domain-aware proxy data $|D_P^*|$, DPS-FD improves model performance and reduces unnecessary communication by selecting representative samples that capture the distribution of each domain.

\begin{figure}[!tp]        
  \centering
  \includegraphics[width=0.9\columnwidth]{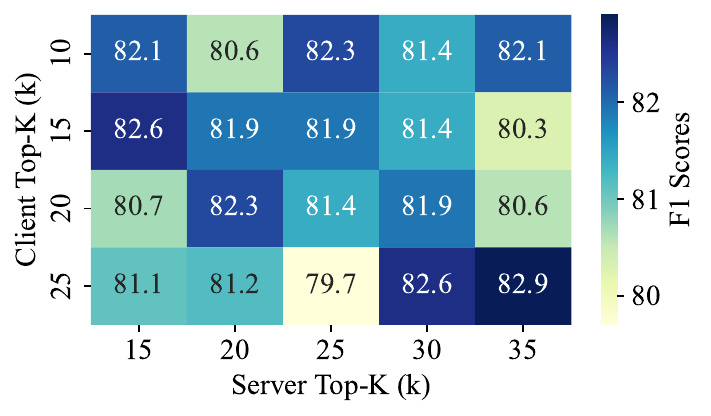} 
  \caption{F1 scores varies with the top-$K$ selections of the clients and server}
  \label{fig:topk-heatmap}
\end{figure}

\subsection{Comparisons with SOTA Models}
As shown in Table~\ref{tab:exit-fd-results}, our proposed DPS-FD significantly outperforms existing FD methods, including DS-FL~\cite{Itahara_2023}, MHAT~\cite{hu2021mhat}, and FedKD~\cite{Wu_2022}, on F1 scores under global distribution performance with proxy data. 
In terms of calibration and uncertainty estimation, DS-FL exhibits poor calibration performance with very low entropy, indicating severe overconfidence in its predictions. In contrast, MHAT and FedKD achieve lower ECE values but produce higher entropy, suggesting more conservative predictions. Our proposed DPS-FD maintains moderate calibration performance with balanced entropy levels, avoiding both excessive overconfidence and overly conservative behavior.
Moreover, DPS-FD achieves this improved performance while reducing communication costs compared to DS-FL and MHAT, benefiting from an effective knowledge exchange process, except for FedKD, which adopts a one-shot distillation paradigm.
This demonstrates superior robustness and generalization across heterogeneous clients, highlighting the effectiveness of domain-aware proxy selection in mitigating distribution shift and improving reliability in FD.

\begin{table}[!tp]
    \vspace{-2mm} 
    \setlength{\tabcolsep}{3pt}  
    \resizebox{\columnwidth}{!}{
    \begin{tabular}{l|c|ccccc|c|c|c}
    \toprule
    Model & \textbf{g.} & \textbf{a.} & \textbf{b.} & \textbf{c.} & \textbf{h.} & \textbf{s.} & cost (\%)  & ECE & entropy  \\
    \midrule
    DS-FL & 80.1 & 79.7 & 82.0 & 78.9 & 78.7 & 80.3 & 100 & 16.74 & 0.12 \\
    MHAT & 81.6 & 79.5 & 83.6 & 80.3 & 72.6 & 78.6 & 100 & 1.66 & 0.44 \\
    FedKD & 82.0 & 80.0 & 84.1 & 82.8 & 73.2 & 78.5 & 16.7 & 2.80 & 0.41 \\
    \midrule
    DPS-FD & \textbf{82.9} & 80.6 & 85.1 & 81.9 & 76.0 & 78.9 & 98.3 & 9.84 & 0.20 \\
    \bottomrule
    \end{tabular}}
    \caption{Results of the global model for existing FD methods under heterogeneous and OOD settings where proxy data is provided.}
    \label{tab:exit-fd-results}
\end{table}

\section{Discussion and Analysis}
We analyze the performance of DPS-FD with respect to the number of clients, top-$K$ selection, proxy data quality, and computational overhead.

\subsection{The Number of Clients}


To investigate the effect of client number, five additional clients from the beauty, patio, pet, shoes, and software domains of the Amazon data are introduced, using \texttt{BERT-base-cased} as the local model.

As shown in Table~\ref{tab:dps_nclient_topk}, global performance consistently improves as the number of participating clients increases, both with and without proxy data. This trend suggests that additional clients provide more diverse domain information, enabling the global model to learn more robust representations under distribution shifts.
Meanwhile, increasing the number of clients generally reduces ECE with slightly lower entropy, indicating more reliable predictions. An exception is observed under the w/ proxy setting with top-$K$ (25k, 35k), where ECE increases when scaling from 5 to 10 clients. This may be due to the enlarged proxy set introducing less informative samples, which slightly degrades calibration reliability.

\subsection{Top-$K$ Selection}
In DPS-FD, the choice of the top-$K$ samples selected by the client and the server plays a crucial role, as it directly affects the trade-off between model performance and communication cost. We vary the number of selected samples on the client side as \{10k, 15k, 20k, 25k\} and on the server side as \{15k, 20k, 25k, 30k, 35k\} under the setting with proxy data.

As shown in Table~\ref{tab:dps_nclient_topk}, with proxy data, the (15k, 15k) configuration achieves a substantial reduction in communication cost (approximately 8.5\% and 3.9\%) while maintaining comparable global performance. Without proxy data, the same (15k, 15k) configuration yields even larger reductions of 18\% and 13.3\%, respectively.
As shown in Figure~\ref{fig:topk-heatmap}, performance does not monotonically improve with smaller top-$K$ values. Smaller top-$K$ retains only highly domain-relevant samples, aggressively filtering
out most hard samples, which may remove informative near-boundary instances and limit generalization. In contrast, larger top-$K$ preserves a portion of such hard samples, which helps improve generalization and decision boundary learning. These results suggest that retaining an appropriate proportion of hard samples is beneficial, rather than strictly filtering them out.

\begin{figure}[t]
    \centering
    \begin{minipage}[b]{\linewidth} 
        \centering
        \includegraphics[width=1\linewidth]{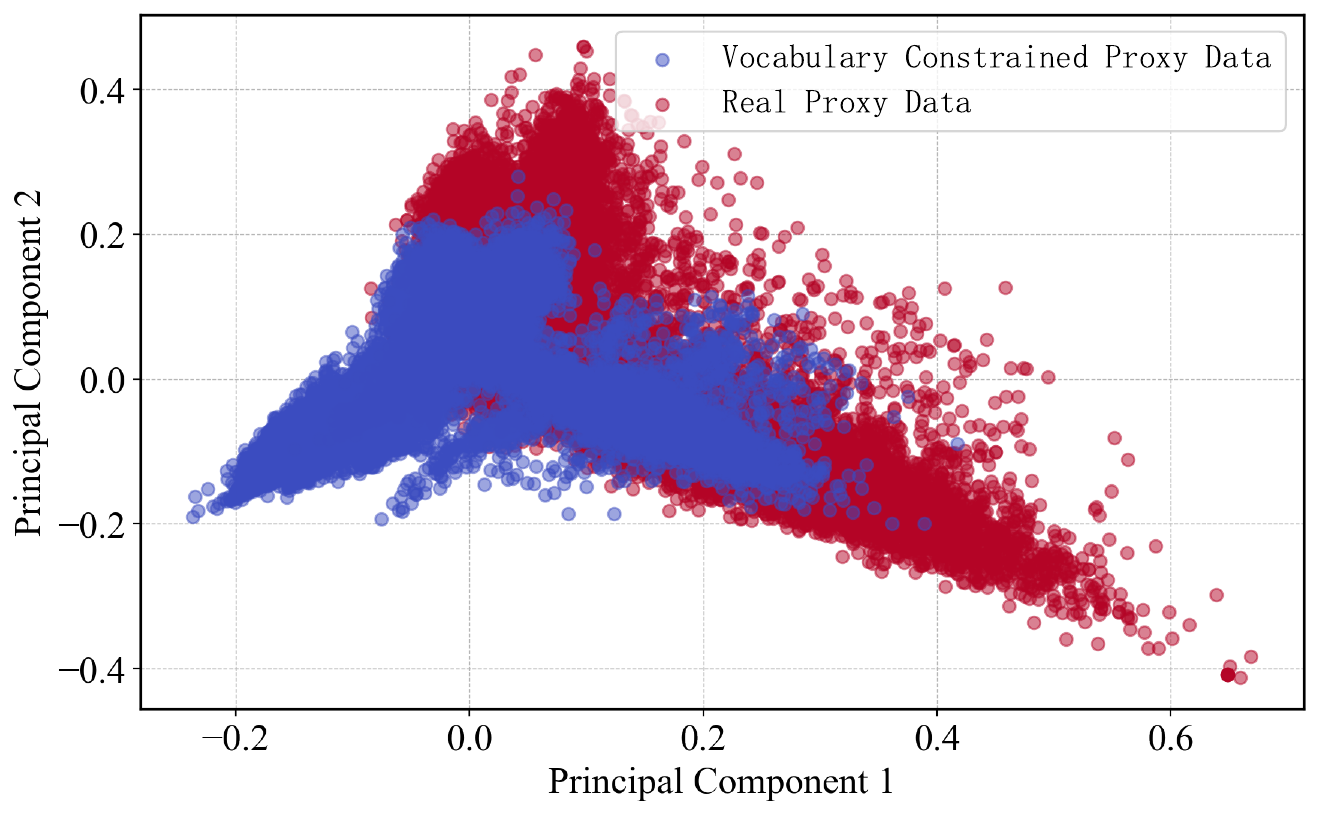}
        \subcaption{Vocabulary-constrained proxy data}
    \end{minipage}
    
    \vspace{0.02\linewidth} 
    
    \begin{minipage}[b]{\linewidth}
        \centering
        \includegraphics[width=1\linewidth]{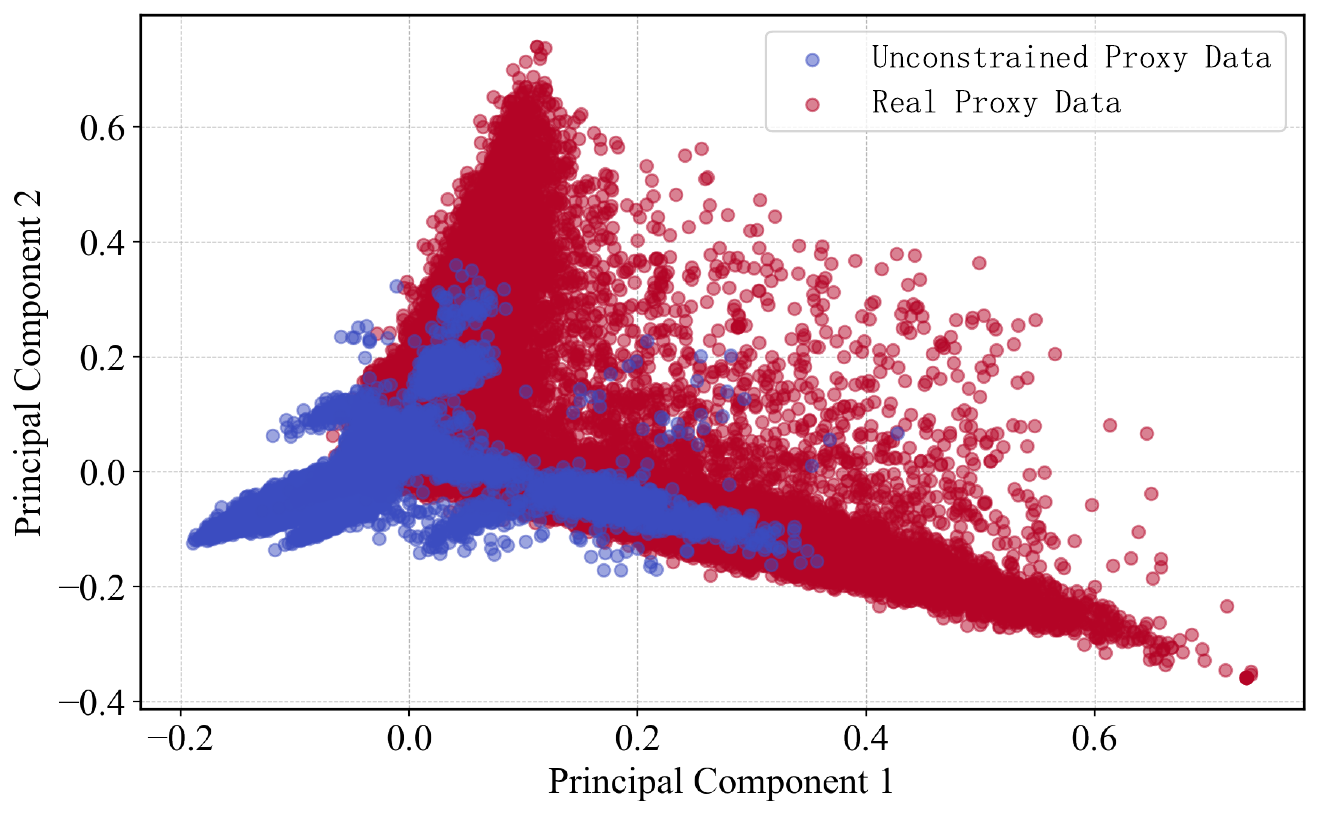}
        \subcaption{Unconstrained proxy data}
    \end{minipage}
    
    \caption{Word distributions of proxy data generated under vocabulary-constrained (top) and unconstrained (bottom) settings, represented in the TF-IDF reduced dimensionality space.}
    \label{fig:proxy-cmp}
\end{figure}

\subsection{Quality of Proxy Data}
We take the insight analysis on vocabulary constraints for the quality of proxy data generated by LLMs. We compare two generation strategies: with vocabulary constraints and without constraints. We tokenize the real data to construct a global vocabulary and enforce the LLM to generate samples using only the words in vocabulary. The unconstrained LLM directly generates samples based on same prompts without vocabulary constraints.

By constraining the vocabulary during LLM-based generation, we guide lexical choices toward distributions closer to real-world data.
This reduces spurious lexical variations and improves semantic consistency of generated samples.
As shown in Figure~\ref{fig:proxy-cmp} (top), proxy data generated with vocabulary constraints is significantly closer to real proxy data, demonstrating the effectiveness of the proposed constraint mechanism.
In contrast, unconstrained generation relies solely on prompt instructions, leading to more diverse but less controlled outputs.
As shown in Figure~\ref{fig:proxy-cmp} (bottom), such samples deviate from real proxy distributions, introducing irrelevant lexical patterns.

Together with the results in Table~\ref{tab:main-results}, these findings indicate that vocabulary constraints help ensure that proxy data remain semantically aligned and more suitable for distillation.

\subsection{Computational overhead}
DPS-FD introduces a lightweight proxy selection step performed locally at each client and the server.
The computational complexity is $\mathcal{O}((\mathcal{K}+1)\cdot |D_P| \cdot d)$, where $d$ is the embedding dimension.
Compared to local training and distillation, which require full forward and backward passes of deep models, the proxy selection only involves cosine similarity computations and thus introduces negligible overhead. Therefore, the size of the proxy data does not become a bottleneck for the scalability of DPS-FD.
The LLM-based proxy data generation is a one-time offline process and does not affect per-round training or communication cost.

\section{Conclusion}
In this paper, 
we propose domain-aware proxy selection that constructs a representative proxy data by jointly considering client-specific and global domain samples to better adopt the proxy data for OOD problems in FD.
Additionally, to address the absence of proxy data, we introduce a vocabulary-constrained LLM-based proxy data generation strategy, which mitigates the generation of linguistically plausible but semantically irrelevant or inconsistent samples. By incorporating the proxy selection strategy with the generation strategy, DPS-FD enhances the robustness and generalization of FD both with and without proxy data. Our experimental results demonstrate that our models outperform existing methods under OOD settings.

\section*{Limitations}
Our method relies on LLMs to synthesize proxy data. LLM-generated data are less reproducible and incur higher API costs and longer response time. We only simulate 5 and 10 clients in our experiments, and we believe that using more clients would be more effective. We take classification as a representative example in this work, and need to conduct more comprehensive validation. We will further explore strategies to reduce ECE while maintaining reasonable uncertainty levels and performance.


\bibliography{acl_latex}
\clearpage

\appendix

\section{Domain Diversity}\label{appendix:domain-diversity}
\begin{figure}[htbp]        
  \centering
  \includegraphics[width=0.9\columnwidth]{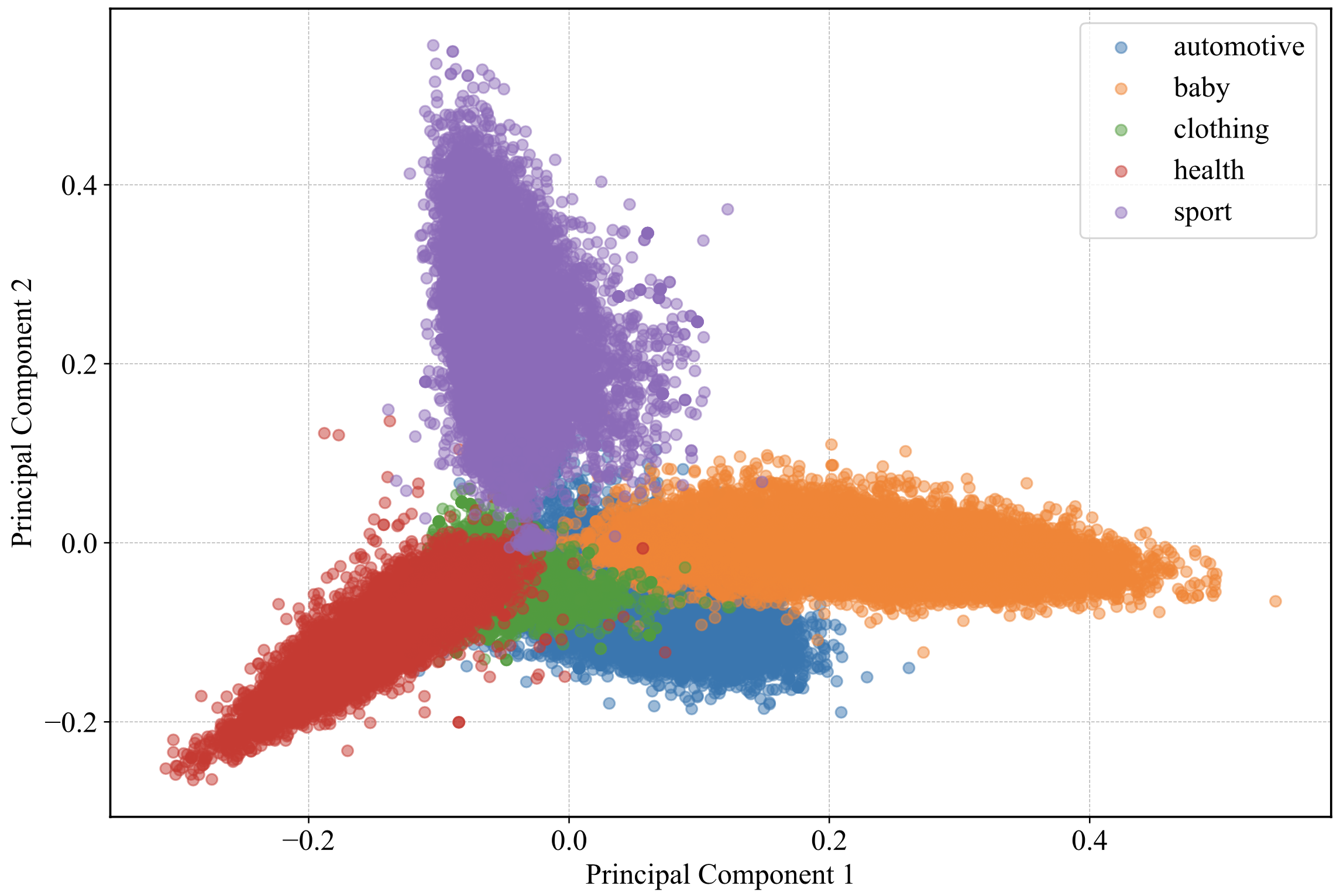} 
  \caption{Domain distributions across different domains.}
  \label{fig:domain-pca}
\end{figure}

To verify that the selected domains exhibit meaningful distribution gaps, we visualize the feature distributions of different domains using Principal Component Analysis (PCA). As shown in Figure~\ref{fig:domain-pca}, samples from different domains show clear separation,
indicating non-trivial distribution shifts among domains. This supports the suitability of the Amazon Reviews dataset for evaluating domain shift and OOD generalization.

\begin{algorithm}[!ht]
    \SetAlgoLined 
	\caption{Domain-Aware Proxy Selection}
        \label{alg:dps-fd}
	\KwIn{labeled private data $\{D_k\}_{k=1}^{\mathcal{K}}$; unlabeled proxy data $D_p$; global model $\theta_g$; local models $\{\theta_k\}_{k=1}^{\mathcal{K}}$; communication rounds $T$}
	\KwOut{global model $\theta_g$}
        \eIf{$D_p$}{
            Use the real proxy data $D_p$ \;
        }{
            Generate a synthetic proxy data $D_p$ using LLM-based generation strategy \;
        }
        \For{each communication round $t=1,2\dots,T$}{
        \textbf{Client executes}:
        
        \For{each client $k$ in parallel}{
            Train local model via \textbf{Eq.}~(\ref{eq:local-train}) \;
            Select and upload the domain-specific proxy data via \textbf{Eq.}~(\ref{eq:centroid})(\ref{eq:proxy-sims})(\ref{eq:proxy-topk}) \;
        }
        
        \textbf{Server executes}: \\
        Select the global domain proxy data via \textbf{Eq.}~(\ref{eq:centroid})(\ref{eq:proxy-sims})(\ref{eq:proxy-topk}) \;
        Construct a domain-aware proxy data $D_p^*$ via \textbf{Eq.}~(\ref{eq:proxy-union}) \;
        Update the global model on $D_p^*$ via \textbf{Eq.}~(\ref{eq:server-distill}) \;

        \textbf{Client executes}: \

            \For{each client $k$ in parallel}{
            Update the local model on $D_p^*$ via \textbf{Eq.}~(\ref{eq:client-distill}) \;
            }
        }
\end{algorithm}
\section{Algorithm of Domain-Aware Proxy Selection}\label{appendix:algo}
Algorithm~\ref{alg:dps-fd} summarizes the overall workflow of DPS-FD and details how the proposed domain-aware proxy selection is incorporated into the FD process.

\begin{tcolorbox}[
  enhanced,
  breakable=false,
  title=System Prompt Template,
  colback=yellow!5,
  colframe=yellow!60!black,
  fonttitle=\bfseries
]
You are an expert language model specializing in generating diverse, high-quality product review texts for federated learning.

You will carefully analyze a set of example reviews and identify corresponding domains and sentiment tendencies.

While the examples may be biased toward certain domains, your task is to summarize them and infer additional potential domains.

Your generation should reflect the authentic style and tone of product reviews: natural, varied, and customer-oriented.

The output must ensure both domain diversity and linguistic diversity, avoiding repetitive templates while maintaining realism.
\end{tcolorbox}

\section{Prompt Templates}\label{appendix:prompt}
The following prompt templates are used to guide the LLM in generating proxy data.
We design both user and system prompts to control the generation process, where the user prompt specifies generation constraints and the system prompt defines overall roles.
These prompts are specifically designed to ensure that the synthesized proxy data can better
approximate real-world review distributions under different domains in federated learning settings.

\begin{tcolorbox}[title=User Prompt Template, colback=blue!5, colframe=blue!60!black, fonttitle=\bfseries, floatplacement=H]

\textbf{A set of representative examples.}

\textbf{Your task:}

1. Identify the product domains and sentiment tendencies represented in the examples above, and infer additional potential domains that could reasonably exist.

2. Based on the examples' and inferred domains, generate exactly realistic and diverse unlabeled review texts.

3. Each review should be fluent, reflecting the natural style of customer reviews.

4. Each review should be about 90 words (minimum 40, maximum 180). Do not shorten because you need to output many.

5. The reviews should not be templates or mechanically repeated. Maintain variability in tone, sentence structure, and vocabulary.

\textbf{Output format:}

1. Output each reviews on a separate line.

2. Ensure the quality of each generated sentence and do not sacrifice quality for the sake of quantity.

3. Do not add any prefix, numbering, headers, or label—only the raw reviews.

Begin now:
\end{tcolorbox}

\end{document}